\documentclass[onecolumn]{jmlr}

\jmlrvolume{}
\firstpageno{1}
\editors{}

\jmlryear{2026}
\jmlrworkshop{NeurReps and WiML Workshops at NeurIPS 2026}

\makeatletter
\def\ps@jmlrtps{\let\@mkboth\@gobbletwo
  \def\@oddhead{\scriptsize\hfill\@jmlrworkshop}%
  \let\@evenhead\@oddhead
  \def\@oddfoot{\@titlefoot}\let\@evenfoot\@oddfoot}
\makeatother

\title[Grid Cells as Spectral Successor Features]{Grid Cells as Spectral Successor
Features:\titlebreak A Reusable Basis for Continual Reinforcement Learning}

\author[C\^ot\'e-Gauthier, Precup and Pr\'emont-Schwarz]{%
 \Name[\'E. C\^ot\'e-Gauthier]{\'Elodie C\^ot\'e-Gauthier}\\
 \addr McGill University and Mila -- Quebec AI Institute
 \AND
 \Name{Doina Precup}\\
 \addr McGill University and Mila -- Quebec AI Institute
 \AND
 \Name{Isabeau Pr\'emont-Schwarz}\\
 \addr Universit\'e Laval and Mila -- Quebec AI Institute
}

\begin{document}

\maketitle

\begin{abstract}
Continual reinforcement learning requires agents to adapt when the task changes while reusing what they know about the environment. Successor features separate dynamics from task-specific reward, but their effectiveness depends on the choice of base features. Taking the identification of entorhinal grid cells with eigenvectors of the successor representation as a design principle, we build this basis from eigenvectors of the transition matrix of a fixed reference policy. With such a basis, the successor features at a state-action pair are a linear combination of the base features at that same pair, which is not true in general, and the coefficients are known in closed form: each successor feature is its base feature multiplied by $\theta_i = 1/(1-\gamma\lambda_i)$, where $\lambda_i$ is its eigenvalue. The long-term return is thus predicted from the same features as the immediate reward, no successor-feature network or replay buffer is needed, and adapting to a new task reduces to a linear regression on the reward. The agent acts greedily on the action values of the reference policy, which is one step of policy improvement over it and suffices to reach a single goal. In a two-goal gridworld, with the uniform random walk as reference policy, whose eigenvectors are the low-frequency Laplacian eigenvectors, agents built this way maintain near-maximal return across every change of goal, whereas SimpleSF and DQN lose most of their return at each change.
\end{abstract}


\section{Introduction}
\label{sec:intro}

Continual reinforcement learning requires an agent to reuse knowledge across tasks when only the reward changes, without relearning the world from scratch. Successor features \citep{barreto2017} address this by decomposing the value function into task-independent dynamics and task-specific reward weights, but their effectiveness depends critically on the feature basis. We ask whether grounding that basis in the spectral geometry of the transition structure, specifically in the eigenvectors of the random-walk graph Laplacian, makes successor-feature learning faster and simpler.

The choice is motivated by neural coding in the hippocampal--entorhinal system. \citet{dayan1993} introduced the successor representation (SR) as a predictive state representation that generalises between states with similar successors, and \citet{stachenfeld2017} proposed that hippocampal place cells encode the SR while entorhinal grid cells correspond to its eigenvectors, so that the SR functions as a predictive map.

\paragraph{Contributions.} (i)~In general, the successor features at a state-action pair cannot be computed from the base features of that pair alone. They can when the span of the base features is invariant under the transition matrix of the policy, as a linear combination with coefficients in closed form, and when the base features are eigenvectors of that matrix, each successor feature is its base feature multiplied by a constant. Successor features then need not be learned, and the long-term return is predicted from the same features as the immediate reward (\sectionref{sec:closed-form}). (ii)~Computing these features for a fixed reference policy, while the agent acts greedily on the action values of that policy, makes the basis independent of the agent's behaviour. The price is that the agent performs one step of policy improvement over the reference policy rather than following the optimal policy, which still suffices to reach a single goal (\sectionref{sec:two-policies}). With the uniform random walk as reference policy, the basis is made of the low-frequency Laplacian eigenvectors, which must be built over state-action pairs, because invariance under the averaged transition operator does not imply invariance under each per-action operator (Sections~\ref{sec:basis} and~\ref{sec:state-action}). (iii)~In a continual-learning gridworld, agents that act greedily on the action values of the uniform random walk, with the transition graph either given or estimated from their own transitions, maintain near-maximal return across changes of goal, where SimpleSF and DQN do not (\sectionref{sec:experiments}).

\subsection{Background}
\label{sec:background}

\paragraph{Setting.} We consider a Markov decision process with finite states $\mathcal{S}$, actions $\mathcal{A}$, transition probabilities $P(s'\mid s,a)$ and discount $\gamma\in(0,1)$. Since the agent needs action values, we work directly with state-action pairs $z=(s,a)$. A policy $\pi$ induces a transition matrix between pairs, $P^\pi_{zz'} = P(s'\mid s,a)\,\pi(a'\mid s')$ for $z'=(s',a')$, and we write $w_j := \mathbb{E}[r(z_j)]$ for the expected immediate reward of pair $z_j$.

\paragraph{The successor representation.} The SR of $\pi$ \citep{dayan1993} counts, for every starting pair, the expected discounted number of future visits to every pair:
\begin{equation}
  M^\pi = \sum_{t=0}^{\infty}\gamma^t (P^\pi)^t = (I-\gamma P^\pi)^{-1},
  \qquad
  Q^\pi(z_i) = \sum_j M^\pi_{ij}\, w_j .
  \label{eq:sr}
\end{equation}
The action value is thus the product of two separate terms: $M^\pi$, which depends only on the dynamics and the policy, and $w$, which depends only on the reward. When the reward changes but the dynamics and the policy do not, $M^\pi$ is kept and $Q^\pi$ is recomputed with a single matrix-vector product.

\paragraph{Successor features.} The SR has one entry per pair of state-action pairs, which does not scale to large or continuous state spaces. Successor features (SFs) \citep{barreto2017} compress it with $n$ base features $\boldsymbol{\phi}(z) = (\phi_1(z),\dots,\phi_n(z))$:
\begin{equation}
  \boldsymbol{\psi}^\pi(z) = \mathbb{E}_\pi\Big[\textstyle\sum_{t=0}^{\infty}\gamma^t \boldsymbol{\phi}(z_t)\ \Big|\ z_0=z\Big] = \textstyle\sum_{t=0}^{\infty}\gamma^t \big((P^\pi)^t\boldsymbol{\phi}\big)(z).
  \label{eq:sf}
\end{equation}
If the reward is linear in the features, $r(z) = \sum_i w_i\phi_i(z)$, then $Q^\pi(z) = \boldsymbol{\psi}^\pi(z)^\top w$. The SR is the special case of one-hot features, for which $w$ is the reward of each pair, as in \equationref{eq:sr}. In general $\boldsymbol{\psi}^\pi$ has to be learned, usually by a neural network trained by temporal difference, and it changes whenever $\pi$ does.

\paragraph{The graph Laplacian and grid cells.} The transitions of the environment also define a graph whose nodes are the pairs, in which $z=(s,a)$ is joined to every pair $(s',a')$ available at the state $s'$ that $a$ leads to. With $W$ its adjacency matrix and $D$ the diagonal matrix of its degrees, $P_{\mathrm{rw}} = D^{-1}W$ is the transition matrix of the uniform random walk on the graph, and $I - P_{\mathrm{rw}}$ its Laplacian. The eigenvectors of this Laplacian with the smallest eigenvalues vary slowly along the graph and describe its large-scale geometry; they are the proto-value functions of \citet{mahadevan2005}. Because the SR of the random walk, $(I-\gamma P_{\mathrm{rw}})^{-1}$, is a function of $P_{\mathrm{rw}}$, it has the same eigenvectors \citep{stachenfeld2017,machado2018}. These shared eigenvectors are the functions that \citet{stachenfeld2017} identify with grid cells.

\subsection{Related work}
\label{sec:related}

Laplacian eigenvectors have also been used in reinforcement learning independently of the neuroscientific connection: as proto-value functions for value approximation \citep{mahadevan2005,mahadevan2007}, for option discovery \citep{machado2017}, and for reward shaping \citep{wu2019}. \citet{machado2018} rely on the same equivalence between the SR and the Laplacian as we do, but in the opposite direction: they learn the SR, from pixels if needed, in order to extract its eigenvectors, which then define intrinsic rewards for exploration. We start from the eigenvectors and derive the successor features from them. \citet{vijayan2026} also learn Laplacian eigenfunctions online from pixels, with an ALLO-based objective \citep{gomez2024} weighted by novelty, and turn them into options that drive exploration towards novel states in sparse-reward Atari and ProcGen games. \citet{touati2023} compare many base features for successor features in zero-shot reinforcement learning and find Laplacian eigenfunctions to be a reliable choice among them, second overall to forward-backward representations, which learn the features and the successor features jointly. They still learn the successor features with a network, for a family of greedy policies, a step our construction removes. Our graph has the topology of the off-policy state-action graph of \citet{osentoski2007}, but we weight its edges differently (\appendixref{apd:weights}) and symmetrise it instead of using a directed Laplacian. They found that Laplacian bases built on this graph approximate action values with far fewer functions than a state basis copied once per action. In their setting the copied state basis remains usable, if less efficient, because Q($\lambda$) learns the effect of each action in the weights. In ours it cannot distinguish actions in general, because the dynamics are fixed in closed form and only the immediate reward is learned, so the action must be carried by the basis itself (\sectionref{sec:state-action}). Finally, SimpleSF \citep{chua2024} learns successor features directly from pixels by combining a temporal-difference loss with a reward-prediction loss, which avoids representation collapse without pretraining. Its network must nonetheless reorganise when the task changes, whereas our successor features are computed rather than learned. We use SimpleSF as our main baseline, together with DQN \citep{mnih2015} as an agent that learns no successor structure.

\section{Spectral successor features}
\label{sec:method}

Two separate choices remove the two problems of \sectionref{sec:background}: that $\boldsymbol{\psi}^\pi$ must be learned, and that it must be relearned whenever the policy changes. Base features whose span is invariant under the transition matrix, such as its eigenvectors, make the successor features closed-form (\sectionref{sec:closed-form}). Computing them for one fixed policy, while the agent acts greedily on the resulting action values, removes the need to recompute them when the agent's behaviour changes (\sectionref{sec:two-policies}).

\subsection{Successor features in the span of the base features}
\label{sec:closed-form}
Following \equationref{eq:sf}, the $i$-th successor feature at the pair $z_t=(s_t,a_t)$ accumulates the $i$-th base feature over the pairs visited from $z_t$ onwards:
\begin{equation}
  \psi_i(z_t) = \mathbb{E}_\pi\Big[\textstyle\sum_{k=0}^{\infty}\gamma^k \phi_i(z_{t+k})\ \Big|\ z_t\Big] = \textstyle\sum_{k=0}^{\infty}\gamma^k \big((P^\pi)^k\phi_i\big)(z_t).
  \label{eq:psi-def}
\end{equation}
It depends on the base features of every pair that can follow $z_t$, not only on those of $z_t$. Seen as vectors over pairs, $\phi_i$ and $\psi_i$ have $|\mathcal{S}||\mathcal{A}|$ entries, one per pair, whereas each pair carries only $n$ base features, usually far fewer. It is therefore not obvious, and not true in general, that the successor features at $z_t$ can be computed from the base features of that same pair,
\begin{equation}
  \psi_i(z_t) = \sum_{j=1}^{n}\alpha_{ij}\,\phi_j(z_t),
  \label{eq:span}
\end{equation}
with coefficients $\alpha_{ij}$ that are the same for every pair, that is, that each $\psi_i$ lies in the span of $\phi_1,\dots,\phi_n$. For a base feature concentrated on a single pair, for instance, $\psi_i$ spreads over every pair from which that pair can be reached, and no combination of a few such features reproduces it. This is why, in general, successor features have to be learned (\sectionref{sec:background}).

\equationref{eq:span} holds exactly when the span of the base features is invariant under the transition matrix, that is, when one step of the dynamics maps each base feature to a combination of base features, $P^\pi\phi_i = \sum_j k_{ij}\phi_j$ for some $n\times n$ matrix $K=(k_{ij})$. Then $(P^\pi)^k\phi_i = \sum_j (K^k)_{ij}\phi_j$, and summing \equationref{eq:psi-def} gives the coefficients in closed form, $\alpha = \sum_k\gamma^k K^k = (I-\gamma K)^{-1}$, a series that converges because the eigenvalues of $K$ are eigenvalues of $P^\pi$, of modulus at most $1$. With a reward linear in the features (\sectionref{sec:background}), the action value $Q^\pi(z_t) = \boldsymbol{\psi}(z_t)^\top w = \sum_{i,j} w_i\,\alpha_{ij}\,\phi_j(z_t)$ is itself linear in the base features of $z_t$: the long-term return, which depends on the transition structure, is predicted from the same local features as the immediate reward, with coefficients computed once.

Eigenvectors of the transition matrix give the simplest invariant span. If $P^\pi\phi_i = \lambda_i\phi_i$, then $K$ and $\alpha$ are diagonal, $(P^\pi)^k\phi_i = \lambda_i^k\phi_i$, and \equationref{eq:span} reduces to
\begin{equation}
  \psi_i(z_t) = \sum_{k=0}^{\infty}\gamma^k\lambda_i^k\,\phi_i(z_t) = \theta_i\,\phi_i(z_t),
  \qquad
  \theta_i = \frac{1}{1-\gamma\lambda_i}.
  \label{eq:psi-i}
\end{equation}
Each successor feature is its own base feature multiplied by a constant; in vector form, $\boldsymbol{\psi}(z) = \Theta\,\boldsymbol{\phi}(z)$ with $\Theta = \mathrm{diag}(\theta_1,\dots,\theta_n)$. The reward and the action value then combine the same features, with the same weights rescaled by $\theta_i$:
\begin{equation}
  r(z) = \sum_{i=1}^{n} w_i\,\phi_i(z),
  \qquad
  Q^\pi(z) = \boldsymbol{\psi}(z)^\top w = \sum_{i=1}^{n} w_i\,\theta_i\,\phi_i(z).
  \label{eq:q}
\end{equation}
The dynamics now enter only through the basis and the constants $\theta_i$, both computed once. The only quantity left to learn is $w$, by a linear regression of the immediate reward on the features, $r(z_t)\approx\boldsymbol{\phi}(z_t)^\top w$, with no successor-feature network, no bootstrapped target and no replay buffer. For an orthonormal basis, collected as the columns of $\Phi\in\mathbb{R}^{|\mathcal{S}||\mathcal{A}|\times n}$, the solution is simply $w^\star = \Phi^\top r$. When the goal moves, only $w$ changes. We call agents built on this decomposition Spectral-SF.

\subsection{Two policies: a fixed reference policy and a greedy acting policy}
\label{sec:two-policies}
The eigenvectors of $P^\pi$ depend on $\pi$, so the basis would have to be recomputed each time the agent's policy changes. We therefore separate the two roles that a policy plays. A \emph{reference policy} $\pi_{\mathrm{ref}}$ is fixed once and for all and is used only for computation: the basis, the constants $\theta_i$ and the action values of \equationref{eq:q} are those of $\pi_{\mathrm{ref}}$. An \emph{acting policy} chooses the actions, greedily with respect to these values:
\begin{equation}
  \pi_{\mathrm{act}}(s) = \arg\max_{a}\, Q^{\pi_{\mathrm{ref}}}(s,a) = \arg\max_{a}\, \sum_{i=1}^{n} w_i\,\theta_i\,\phi_i(s,a).
  \label{eq:pi-act}
\end{equation}
The acting policy changes continuously, as $w$ is learned and whenever the goal moves, but nothing has to be recomputed when it does. The reference policy, on the other hand, can be any fixed policy: the uniform random walk, whose transition matrix depends only on the graph of the environment (\sectionref{sec:background}), a policy learned on an earlier task, or the agent's own exploration averaged over time (\sectionref{sec:online}). \sectionref{sec:agents} gives the choice we make in our experiments.

This separation has a price. \equationref{eq:q} gives the value of $\pi_{\mathrm{ref}}$, not of the acting policy, and the agent acts greedily on it. This is one step of policy improvement over $\pi_{\mathrm{ref}}$, so $\pi_{\mathrm{act}}$ is at least as good as $\pi_{\mathrm{ref}}$, but not optimal in general. With a single positive reward, such as a goal, it is nonetheless enough to reach the goal from every state from which $\pi_{\mathrm{ref}}$ can reach it. For exact values and deterministic transitions, $Q^{\pi_{\mathrm{ref}}}(s,a) = r(s,a) + \gamma V(s'_a)$, where $V$ is the value of $\pi_{\mathrm{ref}}$ and $s'_a$ the state that $a$ leads to, and $V(s) = \sum_a \pi_{\mathrm{ref}}(a\mid s)\,Q^{\pi_{\mathrm{ref}}}(s,a) \le \max_a Q^{\pi_{\mathrm{ref}}}(s,a)$. The greedy action therefore either reaches the goal or leads to a state $s'$ with $\gamma V(s')\ge V(s)$: since $V$ is bounded, it cannot grow by a factor $1/\gamma$ at every step indefinitely, so the goal is reached from every state where $V>0$, that is, from which $\pi_{\mathrm{ref}}$ can reach the goal. When rewards of both signs coexist, this guarantee is lost and the greedy agent can be trapped; \sectionref{sec:limitations} gives an example.

\subsection{The basis of the reference policy}
\label{sec:basis}
The basis consists of eigenvectors of $P^{\pi_{\mathrm{ref}}}$. This matrix is in general not symmetric, so its eigenvectors need not be real or orthogonal. We therefore symmetrise it \citep{wu2019}, weighting each pair by the frequency $\rho$ with which $\pi_{\mathrm{ref}}$ visits it:
\begin{equation}
  L = I - \tfrac12\Big(D_\rho^{1/2}P^{\pi_{\mathrm{ref}}}D_\rho^{-1/2} + D_\rho^{-1/2}\big(P^{\pi_{\mathrm{ref}}}\big)^{\!\top} D_\rho^{1/2}\Big),
  \qquad D_\rho = \mathrm{diag}(\rho).
  \label{eq:sym-laplacian}
\end{equation}
$L$ is symmetric, with real orthonormal eigenvectors $u_i$ and eigenvalues $\mu_i$. When $\rho$ is the stationary distribution of $\pi_{\mathrm{ref}}$, the functions $\phi_i = D_\rho^{-1/2}u_i$ are eigenvectors of $\bar{P} = \tfrac12\big(P^{\pi_{\mathrm{ref}}} + D_\rho^{-1}(P^{\pi_{\mathrm{ref}}})^\top D_\rho\big)$, the walk that follows the transitions of $\pi_{\mathrm{ref}}$ forwards or backwards in time with equal probability, and \equationref{eq:psi-i} gives the successor features of this symmetrised walk exactly, and those of $\pi_{\mathrm{ref}}$ approximately. We keep the $n$ eigenvectors of smallest eigenvalue, which vary most slowly along the graph:
\begin{equation}
  L u_i = \mu_i u_i,
  \qquad
  \bar{P}\phi_i = (1-\mu_i)\,\phi_i,
  \qquad
  \theta_i = \frac{1}{1-\gamma(1-\mu_i)} .
  \label{eq:eigen}
\end{equation}
For the uniform random walk, with transition matrix $P_{\mathrm{rw}}$, the basis is made of the low-frequency eigenvectors of the graph Laplacian of \sectionref{sec:background}, symmetrised as in \equationref{eq:sym-laplacian}: the grid cells.

\subsection{The basis must be built over state-action pairs}
\label{sec:state-action}
A state-only basis, with the action dependence carried by one reward column per action, fails for a geometric reason, which we show for the uniform random walk as reference policy. Write $P_a\in\mathbb{R}^{|\mathcal{S}|\times|\mathcal{S}|}$ for the transition matrix of action $a$, with $(P_a)_{ss'} = P(s'\mid s,a)$, and $P_{\mathcal{S}} = |\mathcal{A}|^{-1}\sum_a P_a$ for their average, the transition matrix of the random walk between states. A state basis $\Phi_{\mathcal{S}}\in\mathbb{R}^{|\mathcal{S}|\times n}$ built from the state-graph Laplacian has a span invariant under $P_{\mathcal{S}}$. Take any function $f$ in this span, and the reward $r = (I-\gamma P_{\mathcal{S}})f$, the same for every action, which is also in the span. The state value of the random walk is then $V = (I-\gamma P_{\mathcal{S}})^{-1}r = f$, and its Bellman equation gives the action values $Q(\cdot,a) = r + \gamma P_a f$, which lie in the span only if $P_a f$ does. Representing the action values of every reward in the span therefore requires the span to be invariant under \emph{each} action, $P_a(\mathrm{span}(\Phi_{\mathcal{S}}))\subseteq\mathrm{span}(\Phi_{\mathcal{S}})$, and not only under their average. Invariance under a sum does not imply invariance under its terms, and there is no reason to expect it here: in a walled environment the $P_a$ do not commute and share no common eigenbasis, since right-then-up differs from up-then-right wherever a wall blocks one order only. Building the graph over state-action nodes, as in \sectionref{sec:background}, makes the action a coordinate of the node, so the basis is built from the operator that the successor recursion iterates, and invariance holds by construction, exactly for $\bar{P}$ and approximately for $P_{\mathrm{rw}}$ (\sectionref{sec:basis}). \appendixref{apd:stateonly} describes a further failure of the state-only parameterisation.

\subsection{Estimating the basis from experience}
\label{sec:online}
The construction above uses the true transitions of $\pi_{\mathrm{ref}}$. The agent can instead build the graph from its own experience. Using the identity of its cell, it counts the transitions it observes: $c(z,z')$ is the number of times it went from the pair $z=(s,a)$ to the pair $z'=(s',a')$, that is, took $a$ in cell $s$, reached $s'$ and then took $a'$. The observed graph has one node per pair, and its empirical adjacency matrix $\hat W\in\mathbb{R}^{|\mathcal{S}||\mathcal{A}|\times|\mathcal{S}||\mathcal{A}|}$ gives the weight $\hat W_{zz'}$ of the edge from $z$ to $z'$: either a binary weight, $\hat W_{zz'} = \mathbf{1}[c(z,z')\ge 1]$, which records only whether the transition has been observed, or the count itself, $\hat W_{zz'} = c(z,z')$. The two weightings correspond to two reference policies (\appendixref{apd:weights}). With binary weights, once every transition has been observed, every next action $a'$ carries the same weight, so the graph is that of the uniform random walk. With counts, each transition is weighted by how often the agent made it, so the graph is that of the agent's exploration policy, averaged over training. A pair that has not been tried yet is treated as an action that leaves the agent in place, so that every node has outgoing edges.

The observed graph is not regular: around the goal, from which the agent never acts, and wherever a pair has not been tried yet, nodes have more incoming than outgoing edges, or the reverse, and with counts the weights also depend on how often each transition was made. \equationref{eq:sym-laplacian} then needs the stationary distribution $\rho$ of the walk on this graph, which is not known in advance. We therefore make the graph undirected, with the symmetric adjacency matrix $\hat A = \hat W + \hat W^\top$, and write $\hat D$ for its degree matrix, the diagonal matrix whose entry $\hat D_{zz} = \sum_{z'}\hat A_{zz'}$ is the total weight of the edges at $z$. The random walk on this undirected graph, whose transition matrix is $\hat D^{-1}\hat A$, has a stationary distribution proportional to the degrees $\hat D_{zz}$, and \equationref{eq:sym-laplacian} applied to this walk gives its normalised Laplacian
\begin{equation}
  \hat L = I - \hat D^{-1/2}\hat A\,\hat D^{-1/2}.
  \label{eq:online-laplacian}
\end{equation}
Let $u_i$ be the eigenvectors of $\hat L$ with the $n$ smallest eigenvalues $\hat\mu_i$. The basis functions are $\phi_i = \hat D^{-1/2}u_i$: they are eigenvectors of the random walk $\hat D^{-1}\hat A$ on the observed graph, with eigenvalue $1-\hat\mu_i$, as \equationref{eq:psi-i} requires, and their $\theta_i$ is given by \equationref{eq:eigen} with $\hat\mu_i$ in place of $\mu_i$. As the agent explores, the observed graph fills in, and the basis approaches that of the reference policy that the weighting defines, except around the goal.

\section{Experiments}
\label{sec:experiments}

\subsection{Environment and task sequence}
We evaluate on the Inverted-LWalls environment of \citet{chua2024}, a $12\times12$ grid with $84$ free cells, without the lava tiles (\sectionref{sec:limitations}) and with four cardinal actions, so that the state-action graph has $336$ nodes. The reward is $1$ on reaching the goal, which ends the episode, and each episode starts from a random cell. Two goal locations, which share the same walls, are presented in sequence and repeated, for $250$k steps each, so that the second exposure to each goal measures retention rather than acquisition. Every agent learns online with $\varepsilon$-greedy exploration, decayed from $1$ to $0.1$ over the first $25$k steps and never reset at a change of goal, and is run with $5$ seeds. \appendixref{apd:setup} gives all hyperparameters.

\subsection{Agents}
\label{sec:agents}
All four agents output the action values $Q(\cdot,a)$ of the four actions and act $\varepsilon$-greedily on them. They differ in what they observe, how they represent the dynamics, and what they learn online (\tableref{tab:agents}).

\begin{table}[!ht]
\floatconts
  {tab:agents}
  {\caption{Inputs, representations and online learning of the four agents.}}
  {\small
  \setlength{\tabcolsep}{4pt}
  \begin{tabular}{lllll}
  \hline
  & DQN & SimpleSF & Offline & Online \\
  & & & Spectral-SF & Spectral-SF \\
  \hline
  Input & pixels & pixels & pixels & cell index \\
  Base features $\boldsymbol{\phi}$ & --- & learned CNN & frozen CNN & eigenvectors of \\
  & & & (pretrained) & observed graph \\
  Successor features & --- & learned by TD & $\Theta\boldsymbol{\phi}$, exact $\theta_i$ & $\Theta\boldsymbol{\phi}$, estimated $\theta_i$ \\
  Learned online & all weights & CNN, $\boldsymbol{\psi}$ head, $w$ & $w$ & graph, $w$ \\
  Replay buffer & $100$k & $100$k & none & none \\
  \hline
  \end{tabular}}
\end{table}

\paragraph{DQN} \citep{mnih2015} receives an $84\times84$ RGB render $x$ of the grid from above. A convolutional network followed by dense layers of $128$ and $256$ units outputs $Q(x,\cdot)$, and all its weights are learned end to end by multi-step double Q-learning from a replay buffer. Geometry and reward are thus entangled in the same weights.

\paragraph{SimpleSF} \citep{chua2024} receives the same render. The same convolutional network outputs normalised base features $\boldsymbol{\phi}(x)\in\mathbb{R}^{128}$, a perceptron head, which also receives $w$, outputs successor features $\boldsymbol{\psi}(x,a)\in\mathbb{R}^{128}$ for each action, and $Q(x,a)=\boldsymbol{\psi}(x,a)^\top w$. The network is trained by a temporal-difference loss on $Q$, and $w$ by a reward-prediction loss, $r \approx \boldsymbol{\phi}^\top w$, both from a replay buffer.

\paragraph{Offline Spectral-SF} receives the same render. Its basis is computed before training: the state-action graph is built from a uniform random walk of $10^6$ steps, long enough to observe every transition, and its $n=128$ eigenvectors of smallest eigenvalue give $\boldsymbol{\phi}$, with $\theta_i$ from the exact eigenvalues (\equationref{eq:eigen}). A convolutional network is then trained offline, by supervised regression on renders of every cell, to output these values, $\boldsymbol{\phi}(x,a)$ for the four actions, from the render $x$, and is then frozen (\appendixref{apd:encoder}). The agent outputs $Q(x,a)=\sum_i w_i\theta_i\phi_i(x,a)$ (\equationref{eq:q}) and learns only $w$, by one gradient step on $(r_{t+1}-\boldsymbol{\phi}(x_t,a_t)^\top w)^2$ after each transition, with no replay buffer.

\paragraph{Online Spectral-SF} receives the index of the agent's cell instead of pixels. It records every transition it observes during training and builds its basis from the observed graph (\sectionref{sec:online}). The basis is first built after $5$k steps and rebuilt every $1$k steps, with $\theta_i$ from the estimated eigenvalues and the sign of each eigenvector aligned with the previous basis, since the eigensolver determines it only up to sign. Its features $\phi_i(s,a)$ are the entries of the eigenvectors, its output and its update of $w$ are those of Offline Spectral-SF, and $w$ is carried over when the basis is rebuilt.

\paragraph{Choice of the reference policy.} Both Spectral-SF variants use the uniform random walk as $\pi_{\mathrm{ref}}$ (\sectionref{sec:two-policies}), so that they act greedily on its action values (\equationref{eq:pi-act}) and their basis is made of the grid cells (\sectionref{sec:basis}). For the online variant, this means binary rather than count weights (\sectionref{sec:online}), the choice that gives the best results in the ablation of \appendixref{apd:weights}. On our graph, away from the goal, every node has as many incoming as outgoing edges (\appendixref{apd:weights}), so $P_{\mathrm{rw}}$ is doubly stochastic and its stationary distribution $\rho$ is uniform. \equationref{eq:sym-laplacian} then reduces to
\begin{equation}
  L = I - \tfrac12\big(P_{\mathrm{rw}} + P_{\mathrm{rw}}^\top\big),
  \label{eq:rw-laplacian}
\end{equation}
$\phi_i$ is proportional to $u_i$, and $\bar{P} = \tfrac12(P_{\mathrm{rw}}+P_{\mathrm{rw}}^\top)$ is itself a random walk, one that follows an edge forwards or backwards with equal probability, whose successor features \equationref{eq:eigen} gives exactly.

In both Spectral-SF variants, $w$ is carried across tasks without reset. \sectionref{sec:limitations} discusses the information each agent is given.

\begin{figure}[!ht]
\floatconts
  {fig:return}
  {\caption{Training return across the task sequence; the goal changes at each quarter of the run. Shaded regions show the standard error over $5$ seeds for each agent.}}
  {\includegraphics[width=0.68\linewidth]{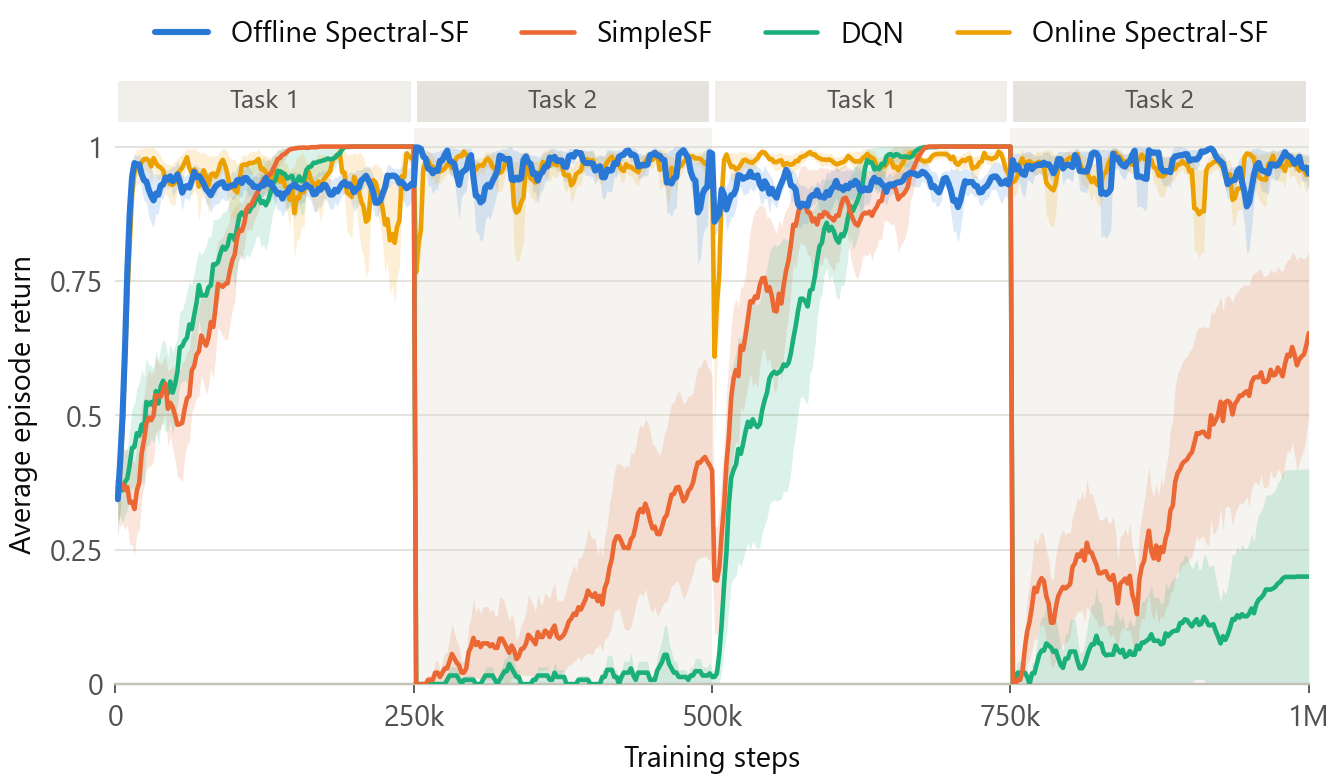}}
\end{figure}

\subsection{Adaptation to changes of goal}
In \figureref{fig:return}, both Offline and Online Spectral-SF approach the maximum return within a small fraction of the first task and remain there across all subsequent changes of goal, with only brief drops at each change, since a change of goal only requires relearning $w$ by linear regression (\sectionref{sec:closed-form}). Both baselines learn the first goal, but their return drops sharply each time the goal moves: SimpleSF must reorganise its features and its value estimate at once, and DQN, which entangles geometry and reward in a single network, has no reusable structure to fall back on. The value field recovered by Spectral-SF is smooth and follows the walls rather than Euclidean distance (\appendixref{apd:vmap}).

Both Spectral-SF variants plateau slightly below the maximum return. This is not the price of acting greedily on the values of the random walk (\sectionref{sec:two-policies}): on this maze, for both goals, acting greedily on the exact action values of the uniform random walk follows a shortest path to the goal from every cell, so that, with exact values, $\pi_{\mathrm{act}}$ is optimal. The gap comes instead from the approximations made in practice: the successor features are those of the symmetrised walk, they are confined to the span of the $n$ retained eigenvectors, which may not represent every reward function with equal fidelity, and $w$ is estimated online. The method therefore trades a slightly lower ceiling for faster adaptation, and in the continual setting of \figureref{fig:return} adaptation dominates.

\section{Discussion}
\label{sec:discussion}

The decomposition is most useful where the geometry of the environment is stable and the reward is not.

\subsection{Limitations}
\label{sec:limitations}
The comparison with SimpleSF and DQN is not on equal information, and should be read as a bound rather than as a head-to-head result. Offline Spectral-SF is an oracle: its basis and $\theta$ come from a random walk that covers every transition (\sectionref{sec:agents}), so it starts with a complete and correct description of the dynamics instead of acquiring it by acting during training, and although it receives the same pixels as the baselines, its encoder is trained offline against this basis on renders of every cell. SimpleSF and DQN, in contrast, must learn their representations from the pixels of their own experience, and adapt them when the task changes. Online Spectral-SF removes the first advantage but not the second: its graph is built from its own transitions, but recording transitions requires the identity of the agent's cell at every step, so it observes the state itself rather than pixels, and its graph is built over the finite set of cells. Both variants therefore measure the value of spectral successor features when the state space is given or directly observed; a fair comparison requires learning the basis from pixels alone.

Because the agent never evaluates its acting policy, no choice of $\pi_{\mathrm{ref}}$ guarantees that this policy is optimal (\sectionref{sec:two-policies}). Consider a mouse separated from a piece of cheese by a narrow corridor whose walls are lava, with a reward of $1$ for the cheese and $-1$ for the lava. The optimal policy walks straight through the corridor. A uniform random walk that enters the corridor, however, is likely to step into the lava before reaching the other end, so the value of $\pi_{\mathrm{ref}}$ is negative in the corridor and in the mouse's room, and decreases towards the corridor: the greedy mouse stays away from it and never reaches the cheese, although the corridor is the only way to it. This is why we removed the lava tiles from our environment, where a single positive reward makes the guarantee of \sectionref{sec:two-policies} hold.

Our evaluation is also limited to simple environments: a single small gridworld with deterministic transitions, a fixed layout and two goal locations. Whether the approach scales to larger, stochastic, continuous or partially observable environments, or to changes of the layout itself, remains to be tested.

\subsection{Future work}
We will remove the privileged information by learning the grid cells online from pixels with the augmented Lagrangian Laplacian objective (ALLO) of \citet{gomez2024}, on which \citet{vijayan2026} already build to learn Laplacian eigenfunctions from pixels in Atari and ProcGen. Trained on the agent's own transitions, it recovers the Laplacian eigenvectors themselves, in order, without the transition graph or the agent's cell, so the agent will observe exactly what the baselines observe. Since the objective does not use the reward, the basis will remain task-independent, and a change of goal will still affect only $w$. We will then evaluate the method in larger and more varied environments.

We will also correct the acting policy, so that the agent is no longer limited to one step of policy improvement over $\pi_{\mathrm{ref}}$, by adding a branch to the network that learns the action values of the acting policy itself, not only those of the fixed reference policy.

\bibliography{archive}

\appendix

\section{Recovered value field}
\label{apd:vmap}

\figureref{fig:vmap} shows $V = \max_a Q(s,a)$ built from the analytic $\theta$ and the learned reward weights on one task. It is smooth, decays with distance from the goal, and respects the walls: cells on opposite sides of a barrier differ sharply despite being spatially adjacent, because the basis encodes distance along the transition graph rather than in the plane.

\begin{figure}[!ht]
\floatconts
  {fig:vmap}
  {\caption{Value map recovered by Offline Spectral-SF on one task, from the analytic $\theta$ and the
   learned reward weights; grey cells are walls. Only the ordering of these values affects
   control, not their scale.}}
  {\includegraphics[width=0.62\linewidth]{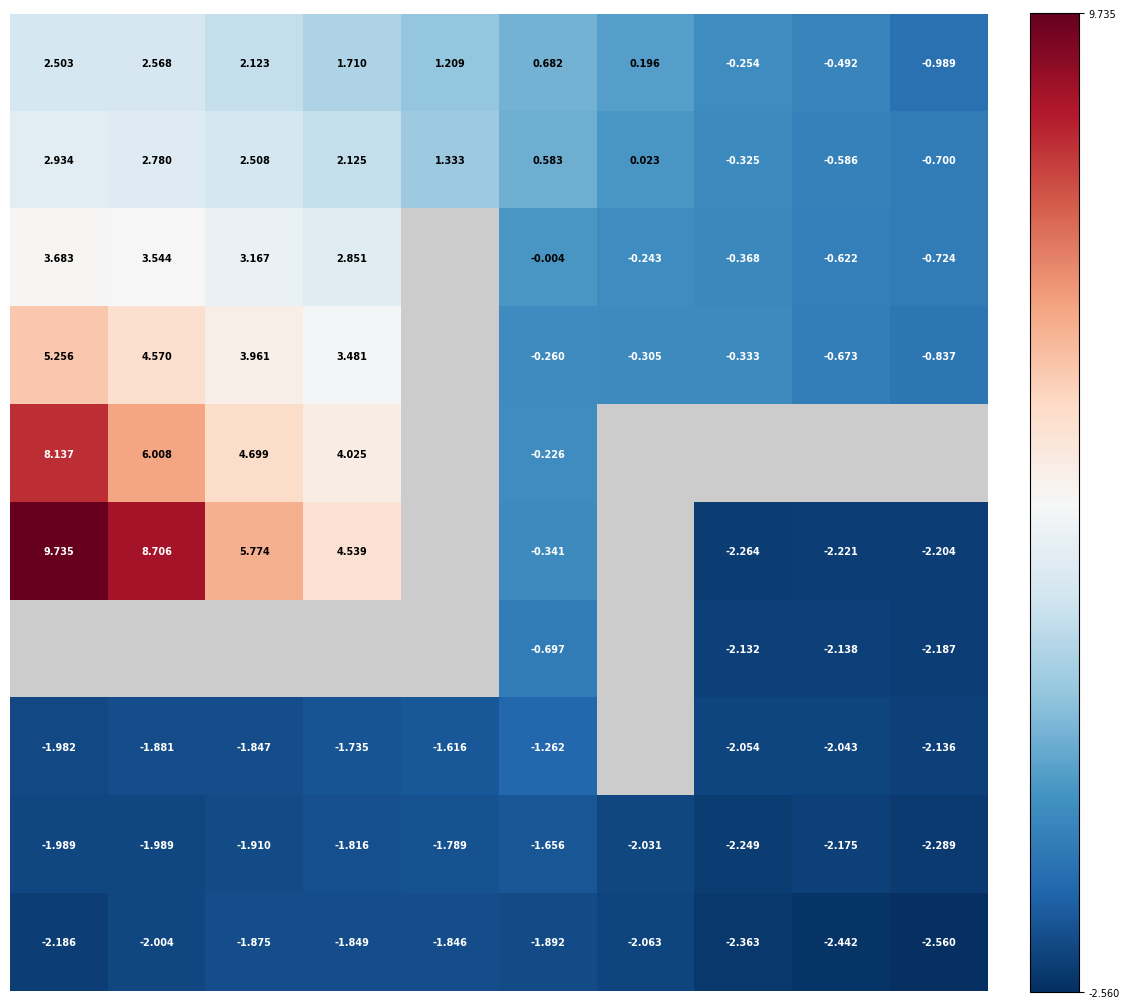}}
\end{figure}
\FloatBarrier

\section{Hyperparameters}
\label{apd:setup}

Episodes are capped at $400$ steps. \tableref{tab:hparams} lists the settings of each agent.

\begin{table}[!ht]
\floatconts
  {tab:hparams}
  {\caption{Hyperparameters of each agent.}}
  {\small
  \begin{tabular}{p{0.22\linewidth}p{0.7\linewidth}}
  \hline
  \multicolumn{2}{l}{\emph{All agents}} \\
  Discount, exploration & $\gamma=0.99$; $\varepsilon=1$ for $5$k steps, then linear to $0.1$ at $25$k steps, never reset \\
  Task sequence & two goals, $250$k steps each, repeated: $1$M steps; $5$ seeds \\
  \hline
  \multicolumn{2}{l}{\emph{DQN}} \\
  Network & CNN (4 convolutional layers, 32 filters of $3\times3$), then dense layers of $128$, $256$ and $4$ units \\
  Optimisation & Adam, learning rate $10^{-4}$; batch $256$; one gradient step every $16$ steps, from $5$k steps into each task \\
  Target & $10$-step double Q-learning; squared TD error with its gradient clipped to $[-10,10]$, equivalent to a Huber loss with threshold $10$; soft target network, $\tau=0.01$, applied every $1000$ steps \\
  Replay & $100$k transitions, emptied at each change of goal \\
  \hline
  \multicolumn{2}{l}{\emph{SimpleSF}} \\
  Network & same CNN, then $128$ normalised base features, a $256$-unit hidden layer that receives them together with $w$, and $128$ successor features per action \\
  Optimisation & network: Adam, learning rate $10^{-4}$; $w$: Adam, learning rate $10^{-5}$, every $16$ steps; target, batch and replay as DQN \\
  \hline
  \multicolumn{2}{l}{\emph{Offline Spectral-SF}} \\
  Basis & $n=128$ eigenvectors of the $336$-node graph, built from a uniform random walk of $10^6$ steps \\
  Encoder & $3$ convolutional and $2$ dense layers, trained offline by $\theta^2$-weighted regression, full batch, $50$k Adam steps, learning rate $10^{-3}$ with cosine decay; then frozen (\appendixref{apd:encoder}) \\
  Reward weights & Adam, learning rate $3\times10^{-3}$, $L_2$ penalty $10^{-3}$, one step per transition \\
  \hline
  \multicolumn{2}{l}{\emph{Online Spectral-SF}} \\
  Basis & binary weights; built after $5$k steps, rebuilt every $1$k steps \\
  Reward weights & as Offline Spectral-SF \\
  \hline
  \end{tabular}}
\end{table}
\FloatBarrier

\section{Training the encoder of Offline Spectral-SF}
\label{apd:encoder}

Offline Spectral-SF observes pixels, but its basis is defined on the cells of the grid. Before reinforcement learning starts, a convolutional network $f$, with three convolutional and two fully connected layers, is therefore trained by supervised regression to output, from the $84\times84$ render $x$ of a cell $s$, the features $\phi_i(s,a)$ of that cell for the four actions, taken from the exact basis of \sectionref{sec:agents}. The targets form a fixed table, and the regression uses no reward. The loss is the squared error weighted by $\theta_i^2$, $\sum_i\theta_i^2\big(f_{a,i}(x)-\phi_i(s,a)\big)^2$, averaged over all renders $x$ and actions $a$, which amounts to regressing the successor features $\Theta\boldsymbol{\phi}$, whose errors are those that enter $Q$ (\equationref{eq:q}). It is minimised with Adam on all renders at once (\tableref{tab:hparams}), down to about $10^{-6}$. The encoder is then frozen, and only $w$ is learned during reinforcement learning.

\section{Ablation: the reference policy, binary versus count weights}
\label{apd:weights}

\sectionref{sec:online} builds the graph of Online Spectral-SF from the counts $c(z,z')$ of the transitions the agent observes, with binary weights, $\hat W_{zz'} = \mathbf{1}[c(z,z')\ge 1]$, or count weights, $\hat W_{zz'} = c(z,z')$, and takes the basis from the normalised Laplacian $\hat L$ of \equationref{eq:online-laplacian}. The two weightings correspond to two choices of the reference policy $\pi_{\mathrm{ref}}$ of \sectionref{sec:two-policies}.

\paragraph{Binary weights: the uniform random walk.} Binarising means keeping only whether a transition has been observed at least once, not how often. Once every transition away from the goal has been observed, every node has $|\mathcal{A}|$ outgoing and $|\mathcal{A}|$ incoming edges of unit weight, hence the same degree $2|\mathcal{A}|$, except around the goal, from which the agent never acts. $\hat L$ then reduces to \equationref{eq:rw-laplacian}: the reference policy is the uniform random walk, whatever the number of visits, and away from the goal the basis is that of Offline Spectral-SF.

\paragraph{Count weights: the average exploration policy.} Normalising the rows of $\hat W$ gives $\hat P$, the empirical transition matrix of the agent between pairs, accumulated over training: $\hat P_{zz'}$ is the fraction of the visits to $z$ after which the agent moved to $z'$, that is, took $a'$ in the cell $s'$ that $z$ leads to. $\hat P$ is thus the transition matrix of the agent's exploration policy averaged over training. With $N$ the number of transitions and $\rho$ the frequency with which each pair was visited, $\hat W = N D_\rho \hat P$ with $D_\rho = \mathrm{diag}(\rho)$. Then $\hat A = N\big(D_\rho \hat P + \hat P^\top D_\rho\big)$ and $\hat D = N\,\mathrm{diag}\big(\rho + \hat P^\top \rho\big)$, and since the factor $N$ cancels in \equationref{eq:online-laplacian},
\begin{equation}
  \hat L = I - D^{-1/2}\big(D_\rho \hat P + \hat P^\top D_\rho\big) D^{-1/2},
  \qquad D = \hat D/N = \mathrm{diag}\big(\rho + \hat P^\top \rho\big).
\end{equation} Up to the restarts of episodes and the goal, $\rho$ is stationary for $\hat P$, so that $D\approx 2D_\rho$ and $\hat L$ is the Laplacian of \equationref{eq:sym-laplacian} with the average exploration policy as $\pi_{\mathrm{ref}}$. It coincides with the binary case when the exploration is uniform. Our agent, however, acts greedily for most of training and restarts each episode from a random cell, so its visits and its choices of action concentrate on the routes to the goal. The count basis then depends on how the agent explored, whereas binary weights recover the environment's graph.

\figureref{fig:binvscount} compares the two weightings on the task sequence of \figureref{fig:return}, with $5$ seeds each. Binary weights, that is, the uniform random walk as reference policy, give the best results.

\begin{figure}[!ht]
\floatconts
  {fig:binvscount}
  {\caption{Training return of Online Spectral-SF using binary weights (uniform random walk as
   reference policy) or count weights (average exploration policy), on the task sequence of
   \figureref{fig:return}. Shaded regions show the standard error over $5$ seeds for each weighting.}}
  {\includegraphics[width=0.68\linewidth]{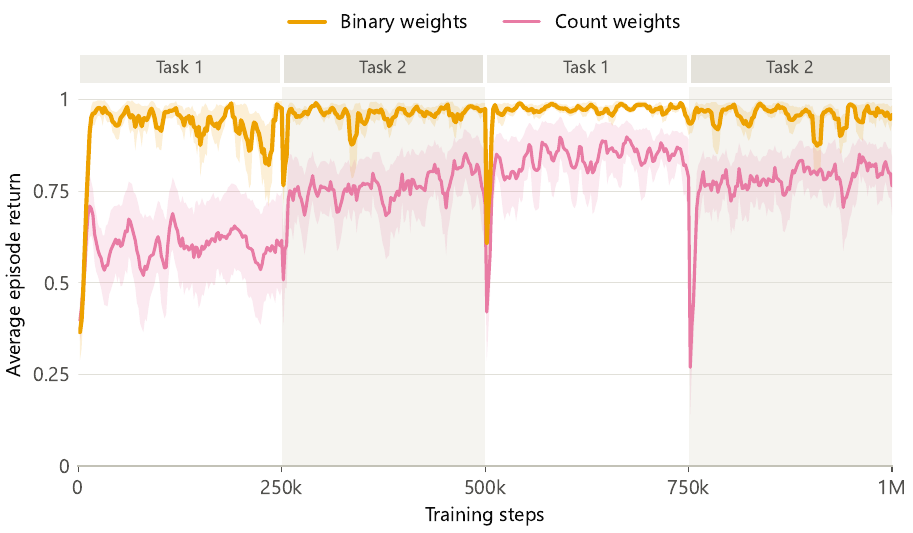}}
\end{figure}
\FloatBarrier

\section{The state-only parameterisation}
\label{apd:stateonly}

With a state-only basis $\boldsymbol{\phi}(s)\in\mathbb{R}^n$, the successor features $\boldsymbol{\psi}(s) = \Theta\boldsymbol{\phi}(s)$ carry no action index, so an action value can only be formed by giving the reward weights one column per action, $w\in\mathbb{R}^{n\times|\mathcal{A}|}$, with $Q(s,a) = \boldsymbol{\psi}(s)^\top w[:,a]$. Beyond the invariance obstruction of \sectionref{sec:state-action}, this parameterisation is fragile in a second way. For an orthonormal basis, whose state features are stacked as the rows of $\Phi_{\mathcal{S}}\in\mathbb{R}^{|\mathcal{S}|\times n}$, the least-squares fit of the immediate reward is performed separately per action, $w[:,a] = \Phi_{\mathcal{S}}^\top r(\cdot,a)$. Writing $s_g$ for the goal and $\mathcal{S}_a = \{s : \text{action } a \text{ at } s \text{ terminates at } s_g\}$, the reward is $r(s,a) = \mathbf{1}[s\in\mathcal{S}_a]$ and
\begin{equation}
  w[:,a] = \Phi_{\mathcal{S}}^\top r(\cdot,a) = \sum_{s\in\mathcal{S}_a}\boldsymbol{\phi}(s).
  \label{eq:so-column}
\end{equation}

\begin{proposition}
If $\mathcal{S}_a = \emptyset$, then $w[:,a] = 0$, and hence $Q(\cdot,a)\equiv 0$.
\end{proposition}

\begin{proof}
The sum in \equationref{eq:so-column} is empty. The same holds when the column is fitted by
gradient descent: the target $r(\cdot,a)$ is then identically zero, and the objective
$\sum_s(\boldsymbol{\phi}(s)^\top w[:,a])^2$ is a positive semidefinite quadratic form minimised at
$w[:,a]=0$, uniquely so under an $L_2$ penalty.
\end{proof}

The premise holds in practice. $|\mathcal{S}_a|$ counts the open neighbours of the goal from which action $a$ steps onto it, so it is fixed by the layout before any learning. A goal walled on two of its four sides leaves two actions with $\mathcal{S}_a = \emptyset$, whose columns are then identically zero; under the greedy rule, such an action is selected only where the remaining action values are non-positive.
\end{document}